%% file: adaptive_dict_kaf.tex
\documentclass[conference]{IEEEtran}
\pdfoutput=1
\ifCLASSINFOpdf
\else
\fi
\usepackage{amsmath}
\usepackage{amssymb}
\usepackage{graphicx}
\usepackage[hyphens]{url} 
\usepackage{epstopdf}
\usepackage{subfig}
\usepackage[font=small,labelfont=bf]{caption}
\usepackage[utf8]{inputenc}
\usepackage{color}

\newcommand{\cH}{\mathcal{H}}
\newcommand{\cD}{\mathcal{D}}
\newcommand{\cN}{\mathcal{N}}
\newcommand{\N}{\mathbb{N}}
\newcommand{\x}{\mathbf{x}}
\newcommand{\balpha}{{\boldsymbol{\alpha}}}
\newcommand{\s}{\mathbf{s}}

\IEEEoverridecommandlockouts

\begin{document}
\title{Initialising Kernel Adaptive Filters \\via Probabilistic Inference}

\author{\IEEEauthorblockN{Iván Castro$^\dagger$\thanks{$^\dagger$These authors contributed equally.}, Cristóbal Silva$^\dagger$}
\IEEEauthorblockA{Department of Electrical Engineering\\
Universidad de Chile\\
\texttt{\{ivan.castro,crsilva\}@ing.uchile.cl}}
\and
\IEEEauthorblockN{Felipe Tobar}
\IEEEauthorblockA{Center for Mathematical Modeling\\
Universidad de Chile\\
\texttt{ftobar@dim.uchile.cl}}
}

\maketitle

\IEEEpeerreviewmaketitle

\input{abstract}
\input{intro}
\input{background}
\input{proposal}
\input{experiments}
\input{conclusions}

\section*{Acknowledgments}
We thank Gonzalo Rios and Angelo Falchetti for useful recommendations of Python's scientific packages, and Conicyt projects PAI-82140061 and Basal-CMM for financial support.
\bibliographystyle{IEEEtran}
\bibliography{library}

\end{document}

%% file: abstract.tex

\begin{abstract}
We present a probabilistic framework for both (i) determining the initial settings of kernel adaptive filters (KAFs) and (ii) constructing fully-adaptive KAFs whereby in addition to weights and dictionaries, kernel parameters are learnt sequentially. This is achieved by formulating the estimator as a probabilistic model and defining dedicated prior distributions over the kernel parameters, weights and dictionary, enforcing desired properties such as sparsity. The model can then be trained using a subset of data to initialise standard KAFs or  updated sequentially each time a new observation becomes available. Due to the nonlinear/non-Gaussian properties of the model, learning and inference is achieved using gradient-based maximum-a-posteriori optimisation and Markov chain Monte Carlo methods, and can be confidently used to compute predictions. The proposed framework was validated on nonlinear time series of both synthetic and real-world nature, where it outperformed standard KAFs in terms of mean square error and the sparsity of the learnt dictionaries.
\end{abstract}

%% file: intro.tex
\section{Introduction}

Within kernel methods, kernel adaptive filters (KAFs) \cite{liu2010} are state-of-the-art nonlinear models for time series that build on the properties of reproducing kernel Hilbert spaces (RKHS) \cite{aronszajn1950theory}, in order to provide accurate predictions at a low computational cost. In the same way that support vectors play a fundamental role in support vector machines \cite{scholkopf01}, KAFs rely on a subset of observed input samples referred to as centres, where new inputs are compared to these centres through a kernel function to compute the prediction. This procedure involves a number of parameters: those of the kernel, those related to the selection of the set of centres (dictionary), and those controlling the trade-off between historical data and new observations. By adapting these model parameters, algorithms, such as kernel least mean square (KLMS) \cite{liu08,richard09} provide an efficient way to improve signal estimation over time as more data become available. Specifically, KLMS applies the least-mean-square rationale to the ``kernelised'' input (i.e., transformed by the kernel function), thus allowing for an efficient online implementation based on gradient steepest descent for updating the model parameters (i.e., the filter weights only).

The main drawback of KAFs is the lack of a principled approach to tune filter weights, kernel parameters and the dictionary. This is mostly due to the fact that the KAF approach is guaranteed to succeed, even when their parameters are not chosen carefully, owing to the universal approximation property of kernels \cite{steinwart01}. However, this may result on suboptimal implementations that, for instance, require large non-sparse dictionaries. Therefore, in our view, a theoretically-grounded parameter setting is required in order to achieve an efficient and accurate operation of KAFs. In \cite{tobar14mk} it was shown that different kernel widths can be used to predict wind-speed signals of different dynamic regimes, thus motivating the search for optimal parameters for the problem at hand; then, \cite{tobar_quat} provides a heuristic rule for parameter setting based on the histograms of the input samples. Another effort for parameter tuning within KAF has been achieved by relating KAFs to Gaussian processes (GP) \cite{rasmussen06}, where a GP-interpretation of the kernel recursive least squares (KRLS) tracker \cite{van2012kernel} allows for a probabilistic interpretation of KAF and therefore training. Additionally, online setting of dictionary and kernel hyperparameters have been studied by \cite{wang2015,chen2016kernel,fan2016kernel} using gradient-based optimisation. Alternatively, a variant of KAFs where parameter setting is straightforward due to preprocessing the observations can be found in \cite{dsp2017b}.

We propose a novel algorithm to completely train a kernel adaptive filter, that is, to find appropriate kernel parameters, weights and dictionary, using a probabilistic interpretation of KAFs. In addition to the probabilistic formulation that allows for training, our main contribution is the design of a sparsity-inducing prior distribution to determine an appropriate dictionary, which unlike standard KAFs is not restricted to be a subset of the observations. Our method can be implemented offline, where a subset of data is used to calculate initial conditions for the parameters of a standard KAF, or online, where a sliding window is used to perform recursive training and prediction.  We validate our approach using both illustrative examples with synthetic data and a real-world wind time series.

The paper is organised as follows: Section II gives a brief overview of KAFs and probabilistic inference; Section III presents the proposed methodology with an example using the Lorenz attractor time series; and Section IV shows the validation of the proposed method and compares it against standard KAFs. Finally, the conclusion and discussion of our findings are presented in Section V.

%% file: background.tex

\section{Kernel Adaptive Filters and Probabilistic Inference}

\subsection{Kernel adaptive filters}

A kernel adaptive filter (KAF) is a nonlinear autoregressive model for time series, whose parameters are updated in an online manner as new data arrive. In detail, for a discrete-time signal given by $\{y_i\}_{i\in\N}$, a KAF aims to predict $y_i$ using past values $y_{i-d:i-1}=[y_{i-d},\ldots, y_{i-1}]$ by first embedding the trajectory $y_{i-d:i-1}$ onto an infinite-dimensional feature space $\cH$ (an RKHS) to then perform a linear estimation. Denoting the input trajectory as $\x_i = y_{i-d:i-1}$, the KAF estimate is then expressed by
\begin{equation}
	\hat{y}_i = \langle {\phi_{\x_i}, W }\rangle
\end{equation}
where $\phi_{\x_i}$ is the element of the RKHS $\cH$ associated to the input sample $\x_i$, $W\in\cH$ is the weight of the defined estimator, and $\langle{\cdot, \cdot}\rangle$ is the inner product in $\cH$.
The above model requires finding an appropriate $W\in\cH$; this is challenging, since $\cH$ is an infinite-dimensional space and it is computationally impossible to implement optimisation routines on this space. Fortunately, the Representer Theorem \cite{scholkopf01} gives a sidestep to this issue, since it states that the optimal $W$ has the form of a sum of evaluations of the map $\phi$ on the observed data. However, this is impractical in online applications where the observations grow unbounded, thus KAFs surmount this issue by choosing a subset of these observations in order to reduce computational complexity and therefore allowing for online operation. The functional form of the weight at time instant $i$, $W_i$ is then given by 
\begin{equation} 
	W_i = \sum_{j=1} ^{N_i} \alpha_{j}\phi_{\s_j}
	\label{eq:weights}
\end{equation}
where $\s_j$ is the $j^\text{th}$ centre, the set $\cD_i = \{\s_j\}_{j=1:N_i}$ is known as dictionary up to time instant $i$, and $\{\alpha_{j}\}_{j=1:N_i}$ are new finite-dimensional weights---we have omitted the explicit dependence of the weights  $\alpha_j$ on time for notational simplicity. Due to the properties of the RKHS \cite{scholkopf01}, the estimator becomes
\begin{equation}
\hat{y}_i = \sum_{j=1}^{N_i} \alpha_j K(x_i, s_j).
\end{equation}
Online training of KAFs then consists in a recursive update of the weights $\{\alpha_{j}\}_{j=1:N_i}$ and the dictionary $\cD_i$. The weights update is usually based on standard linear filters (since the model is still linear in the parameters $\{\alpha_{j}\}_{j=1:N_i}$), such as least mean square \cite{liu08,richard09}, recursive least squares \cite{engel04}, state-space models \cite{tobar15a}. The dictionary learning is referred to the sparsification criteria \cite{honeine15} and addresses the trade off between the magnitude of the prediction error and the relative distance between input samples and dictionary centres. Popular sparsification criteria for KAFs include the coherence \cite{richard09} and novelty \cite{platt91} criterion. 

For Gaussian kernels, dictionary selection is closely related to the kernel lengthscale, since changing the lengthscale completely varies the sparsity and similarity of the dictionary. Both dictionary learning and lengthscale setting are rarely addressed as a whole, but they are usually approached independently. Current sparsification methods only address the dictionary construction, even though it is clear that the kernel parameter impacts whether or not data and dictionary ``look alike''. Other approaches to define the best fit rely on cross-validation and hand-picked parameters such as Silverman's Rule \cite{silverman1986density}.

\subsection{Probabilistic inference} 
\label{sub:inference}
Probabilistic inference allows for representing uncertainty, in this context, a model is expressed as $p(y|x,\theta)$, where $y$ is the output, $x$ the input and $\theta$ the model parameters. To find the parameters given observed input-output pairs we can use Bayes theorem, which requires to define a \textit{prior} distribution on the parameter $\theta$, where then the \textit{posterior} distribution can be either maximised or calculated (approximated) to perform predictions. The probabilistic standpoint provides a principled approach for finding model parameters and has not been used in the KAF context, our hypothesis is that combining these concepts opens new avenues to train KAFs as we will see in the remaining of the paper.

%% file: proposal.tex

\section{A Probabilistic View of Kernel Adaptive Filters} 
\label{sec:a_novel_kernel_adaptive_filter}

Recall that standard KAF methods approximate the optimal weight element $W_i$ in eq. \eqref{eq:weights} by a subset of feature transformations of the observations. Yet simple, this is approach is rudimentary when compared to other sparsification procedures (see e.g. sparse GPs \cite{quinonero2005unifying}, where the centres are referred to as \emph{inducing points}). In fact, it is known that optimising over the centres/inducing points is more appropriate \cite{tobar_npk,zoubin_sparse}. Thus, we propose a probabilistic formulation of KAFs that allows for a principled choice of the dictionary, its initial weights, the kernel parameters and the noise variance.

\subsection{Generative Model}
Rather than using a suboptimal approximation based on the Representer Theorem, we define the dictionary $\cD_i=\{\s_j\}_{1:N_i}$, as well as the weights $\{\alpha_j\}_{1:N_i}$ and the remaining  hyperparameters in a probabilistic manner. This definition preserves the original formula of the KAF estimator with an added observation noise term, that is, 

\begin{equation}
y_i = \sum_{j=1}^{N_i} \alpha_j K_{\sigma_k}(\x_i, \s_j) + \epsilon_i
\label{estimator}
\end{equation}
where $\epsilon_i\sim\cN(0,\sigma_\epsilon^2)$, $K_{\sigma_k}$ is a kernel with parameter $\sigma_k$, and we denote the input at time $i$ as $\x_i = [y_{i-d},\ldots, y_{i-1}]$, choosing a model order equal to $d$ without loss of generality. We emphasise that all the quantities are random variables and not chosen from heuristics as in the standard KAF setting, in particular, the dictionary $\cD=\{\s_1,\ldots,\s_{N_i}\}$ is not necessarily a subset of the observed inputs $\x_i,\ i\in\N$. We now focus on fixed dictionaries and have denoted $\cD_i=\cD$.

For an observed trajectory $Y=\{y_1,\ldots, y_N\}$ the model likelihood can be written in product form, since the process $y_i,\ i\in\N$, is $d$-order Markovian and admits the decomposition
\begin{equation}
\begin{split}
    p(Y) &= \prod_{i=d}^{N} p(y_i\vert \x_i) \\ &= \prod_{i=d}^{N} \frac{1}{\sqrt{2\pi\sigma_\epsilon^2}}\exp\left( -\frac{\left(y_i - \balpha^\top K_{\sigma_k}(\x_i, \cD)\right)^2 }{2\sigma_\epsilon^2}\right) 
\end{split}
\end{equation}
where $\balpha_i=[\alpha_1,\cdots,\alpha_{N_i}]^\top$ is the vector of filter weights at time $i$, and $K_{\sigma_k}(\x_i, \cD)$ denotes the vector of kernel evaluations of the input $\x_{i}$ and each element of the dictionary $\cD$. Furthermore, we can place priors on the weight vector $\balpha_i$, the kernel parameter $\sigma_{k}$ and the noise variance $\sigma_\epsilon$:
\begin{align*}
    p(\balpha) &= \frac{1}{\sqrt{2 \pi l_\alpha^2}}\exp\left( -\frac{\left\| \balpha \right\|^2}{2 l_\alpha^2}  \right)
    \label{eq:coef_prior}\\
   	p(\sigma_{k}) & = \mathcal{N}_{\mathbb{R}^+}(0,v_k)\\
	p(\sigma_{\epsilon}) & = \mathcal{N}_{\mathbb{R}^+}(0,v_\epsilon)
\end{align*}
where $p(\balpha)$ enforces small and regular filter weights, and the half-Normal priors on both the kernel parameter $\sigma_k$ and the noise variance $\sigma_\epsilon$ ensure positive and close-to-zero values. 

We propose to use the following prior for the dictionary
\begin{equation}
	 p(\cD) = \frac{1}{\sqrt{2 \pi l_\cD^2}}\exp\left( -\frac{\left\|K_\sigma(\cD,\cD) \right\|^2}{2 l_\cD^2}  \right)
	\label{eq:dict_prior}
\end{equation}
this is an exponential distribution on the square norm of the Gram matrix evaluated on the dictionary. Consequently, this prior produces dictionaries for which $\left\|K_\sigma(\cD,\cD) \right\|^2$ is close to zero. For a Gaussian kernel, this implies that the samples of the dictionary will be far from one another, since $\left\|K_\sigma(\cD,\cD) \right\|^2 = \sum_{j,k=1}^{N_i}K^2_{\sigma_k}(\s_j,\s_k)$. Therefore, $p(\cD)$ in eq. \eqref{eq:dict_prior} is a sparsity-inducing prior that avoids redundancy arising from choosing the centres directly from data, as it is the case in the coherence \cite{richard09} or novelty \cite{platt91} sparsification criteria. Furthermore, we place half-Normal hyperpriors on $l_\alpha$ and $l_\cD$, with zero mean an variances $v_{\balpha}$ and $v_{\cD}$ respectively.

Although the full posterior can be calculated analytically (up to a normalising constant), we do not show it here due to space constraints. However, observe that besides the regularisation terms, the three main terms of the log-posterior are
\begin{equation*}
\begin{split}
  \frac{-\sum_{i=1}^{N} \left(y_i - \balpha^\top K_{\sigma_k}(\x_i, \cD)\right)^2}{2\sigma_\epsilon^2} -\frac{\left\|K_\sigma(\cD,\cD) \right\|^2}{2 l_\cD^2} -\frac{\left\|\balpha \right\|^2}{2 l_\balpha^2}.
\end{split}
\label{eq:log-posterior}
\end{equation*}
Therefore, optimisation of the log-posterior of the proposed model is a trade-off among data fit (first term), sparsity of the dictionary (second term) and regular weights (third term).

\subsection{Offline optimisation of the log-posterior}
 
To illustrate the suitability of the proposed approach to find appropriate parameters, we present the following synthetic-data example. We generated 1000 observations of the first channel of the Lorenz Chaotic Attractor \cite{lorenz1963deterministic} and implemented a kernel estimator as that in eq. \eqref{estimator}, with dictionary of size 5 and order $d=5$. We considered different subsets for training and then predicted the remaining part of the time series with the (fixed) parameters learned according to the model in eq. \eqref{estimator}. The motivation for this experiment is to assess how the predictive ability of the  model improves as more data are seen. 

\begin{figure}[t!]
	\centering
	\includegraphics[width=0.5\textwidth]{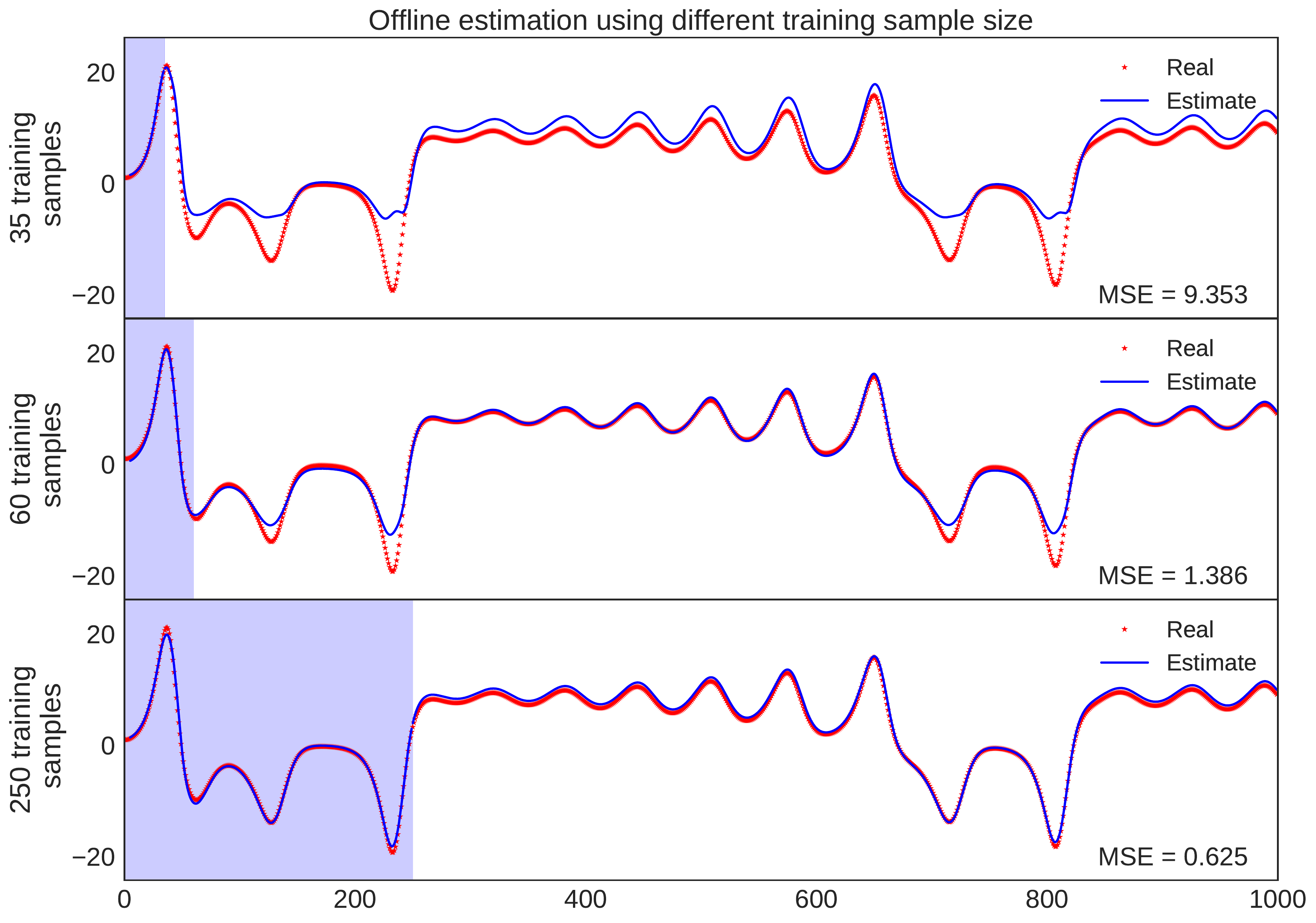}
	\caption{Prediction of the Lorenz series using the model in eq. \eqref{estimator} with fixed parameters trained with different subsets of data. The blue area indicates the training data in each case.}
	\label{fig:lorenz-series}
\end{figure}

\begin{figure}[t!]
	\centering
	\includegraphics[width=0.2\textwidth]{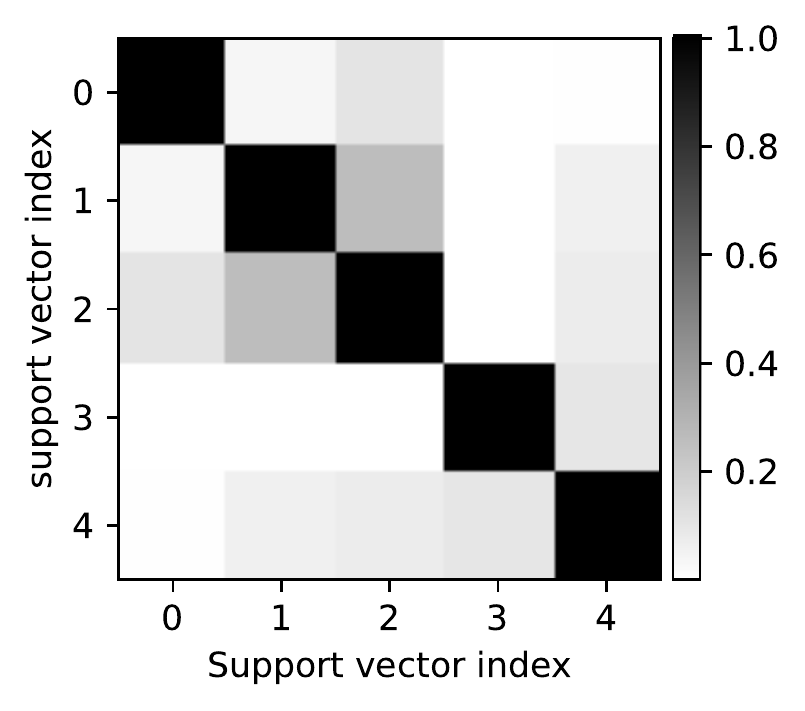}
	\caption{Gram matrix of size $5$ for the $250$ sample training case.}
	\label{fig:lorenz-gram}
\end{figure}

Fig. \ref{fig:lorenz-series} shows the prediction of the model trained using 35 (top), 60 (middle) and 250  (bottom) samples, where the chosen parameters were set to the mean of the sample posterior approximated using Markov chain Monte Carlo (MCMC). Notice how the estimation improves pointwise, reaching all the extrema of the signal for the 250-training-sample case, even though this is a non-adaptive model. Fig. \ref{fig:lorenz-gram} shows the Gram matrix for the optimal centres for the case of 250 training samples, just as expected, sampling from eq. \eqref{eq:dict_prior} delivers a sparse dictionary characterised by a close-to-diagonal Gram matrix.

%% file: experiments.tex

\section{Simulations: Online Operation}

We envision two practical uses of the proposed approach for online time-series prediction: The first one is to determine initial settings for a KAF algorithm (e.g., KLMS), and the second one is to implement the proposed algorithm online using a sliding window. We now show experimental results for these cases using a real-world wind signal from the 2011 PHM Society Conference Data Challenge \cite{windref} and relied on the Python toolbox PyMC3 for maximising or approximating the posterior density.

\subsection{Pre-training for kernel adaptive filters}

We first implemented the proposed method to provide initial conditions for the dictionary, weights, kernel parameter and noise variance for a KLMS algorithm. This aimed to avoid the usual hand-tuning design of KAFs---we refer to this procedure as \emph{pre-training}. The wind data had 2500 samples where the first $270$ samples were used to train our method; we then implemented two KLMS predictors: one with the parameters determined using a maximum-a-posteriori (MAP) fit of the proposed model with fixed dictionary after training, and a standard KLMS with novelty sparsification criterion. For standard KLMS, the novelty criterion parameters were set to have the same number of centres as the proposed model; this is to validate the ability of our method to generate sparse dictionaries against the KLMS, which simply populates the dictionary from the observations.

In both cases the order of the filter was set to $d=5$ and the learning rate was adjusted in each experiment so as to minimise the mean square error (MSE), defined by
\begin{equation*}
\text{MSE} = \frac{1}{n}\sum_{i=1}^{N}\left(y_i - \hat{y}_i\right)^2
\end{equation*}
where $\hat{y}_i$ corresponds to the prediction of the observation $y_i$. 

\begin{figure}[t!]
	\centering
	\subfloat[]{\includegraphics[width=0.165\textwidth]{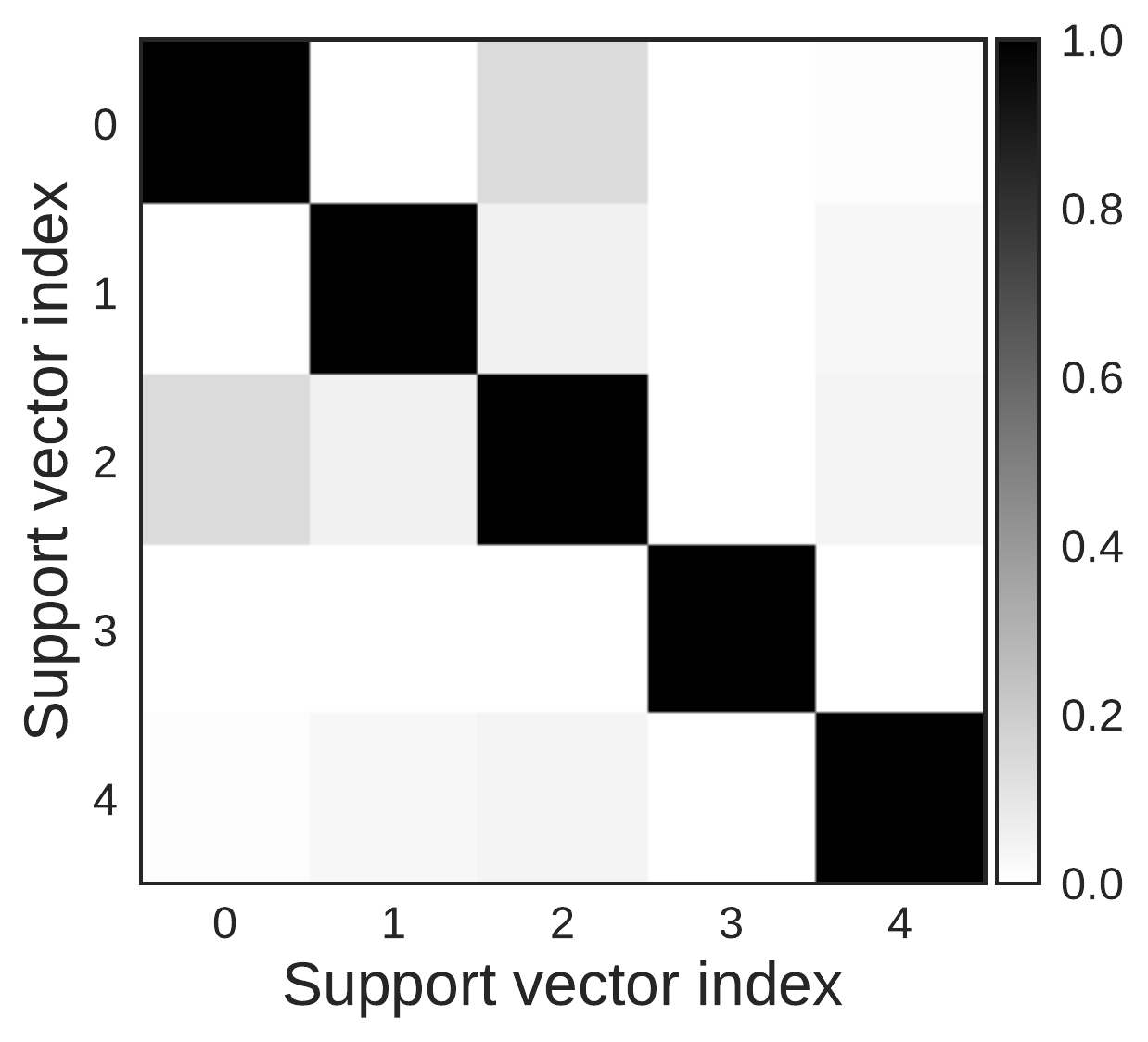}}
	\subfloat[]{\includegraphics[width=0.17\textwidth]{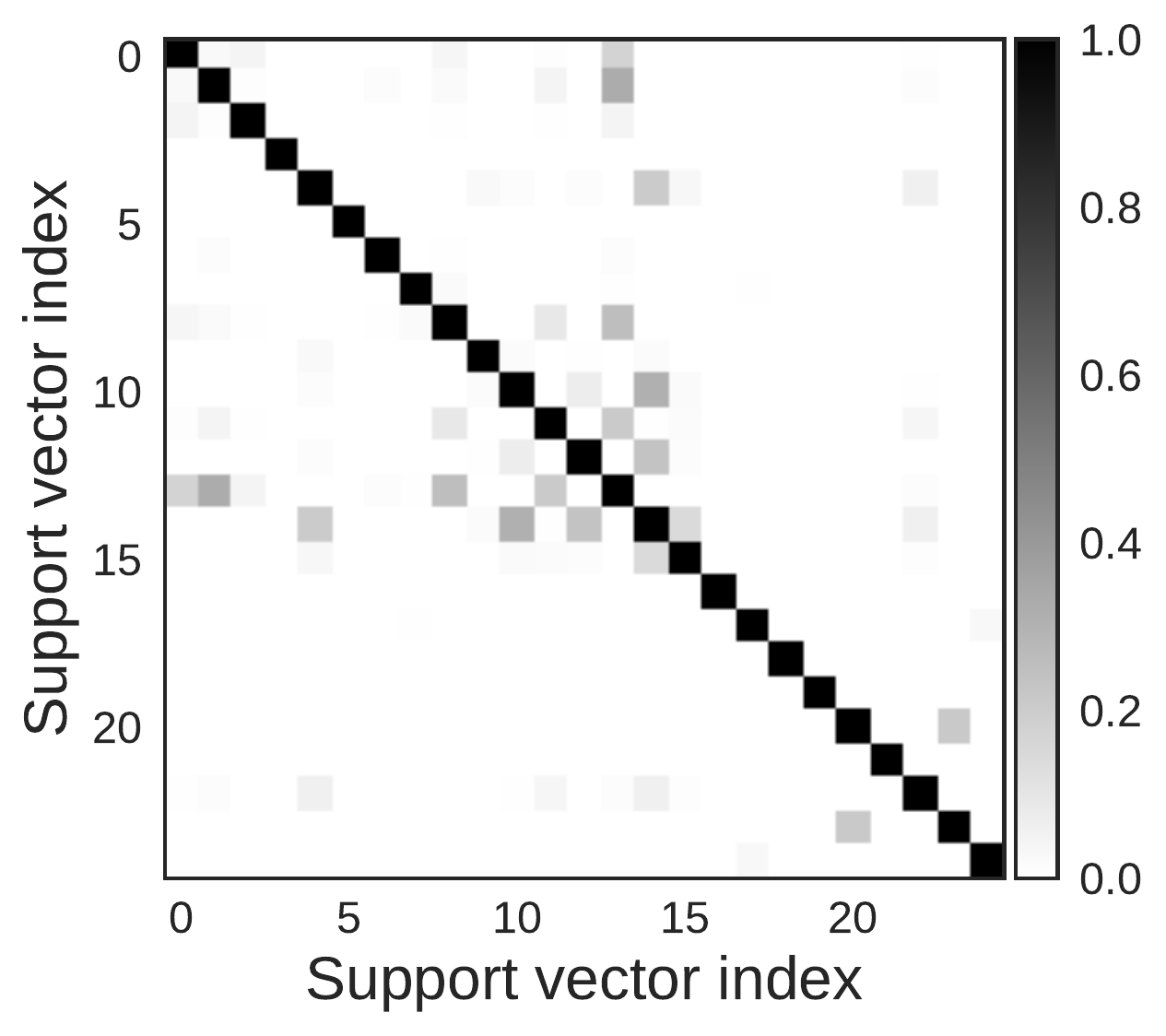}}
	\subfloat[]{\includegraphics[width=0.17\textwidth]{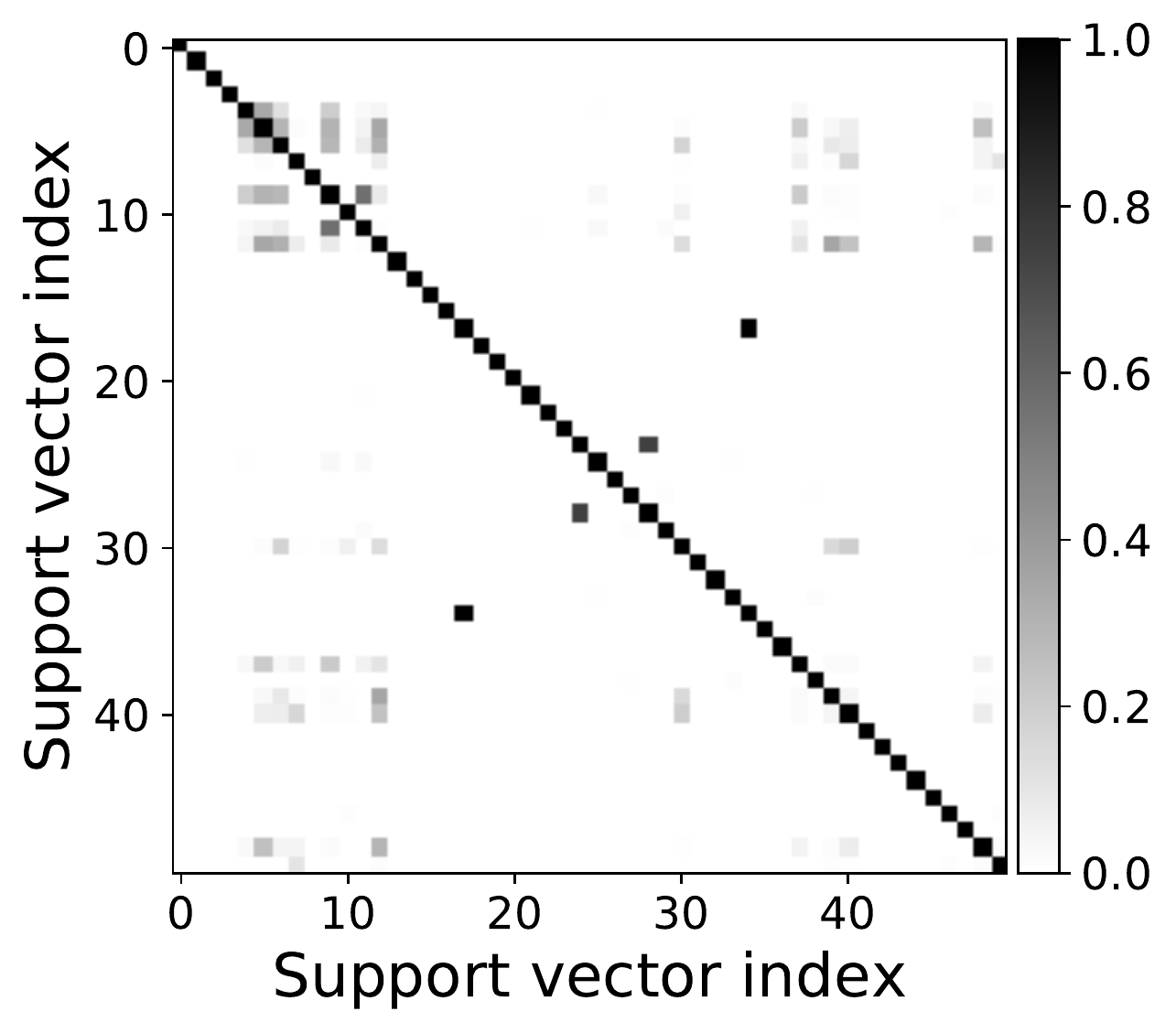}}
	\caption{Gram matrices of pre-trained KLMS for wind signals using dictionaries of sizes 5, 25 and 50 (from left to right). In all cases the Gram matrix is close-to-diagonal due to the sparsity-inducing prior.}
	\label{wind-confmats}
\end{figure}

In order to set the number of centres for pre-training KLMS, we compared different dictionary sizes to assess whether sparsity is achieved independent from the dictionary size. Fig. \ref{wind-confmats} shows the Gram matrices after optimisation of the log-posterior for different choices of dictionary sizes, where it can be seen that the method produces sparse dictionaries for all these choices. We then chose $25$ centres to ensure a sufficiently-rich dictionary, as well as to avoid the increased computational complexity associated by larger dictionaries. Notice that this is the opposite to what we usually do in standard KLMS, where more dictionary elements are needed to compensate for the redundant information that exists among the centres.

\begin{figure}[t!]
	\centering
	\includegraphics[width=0.5\textwidth]{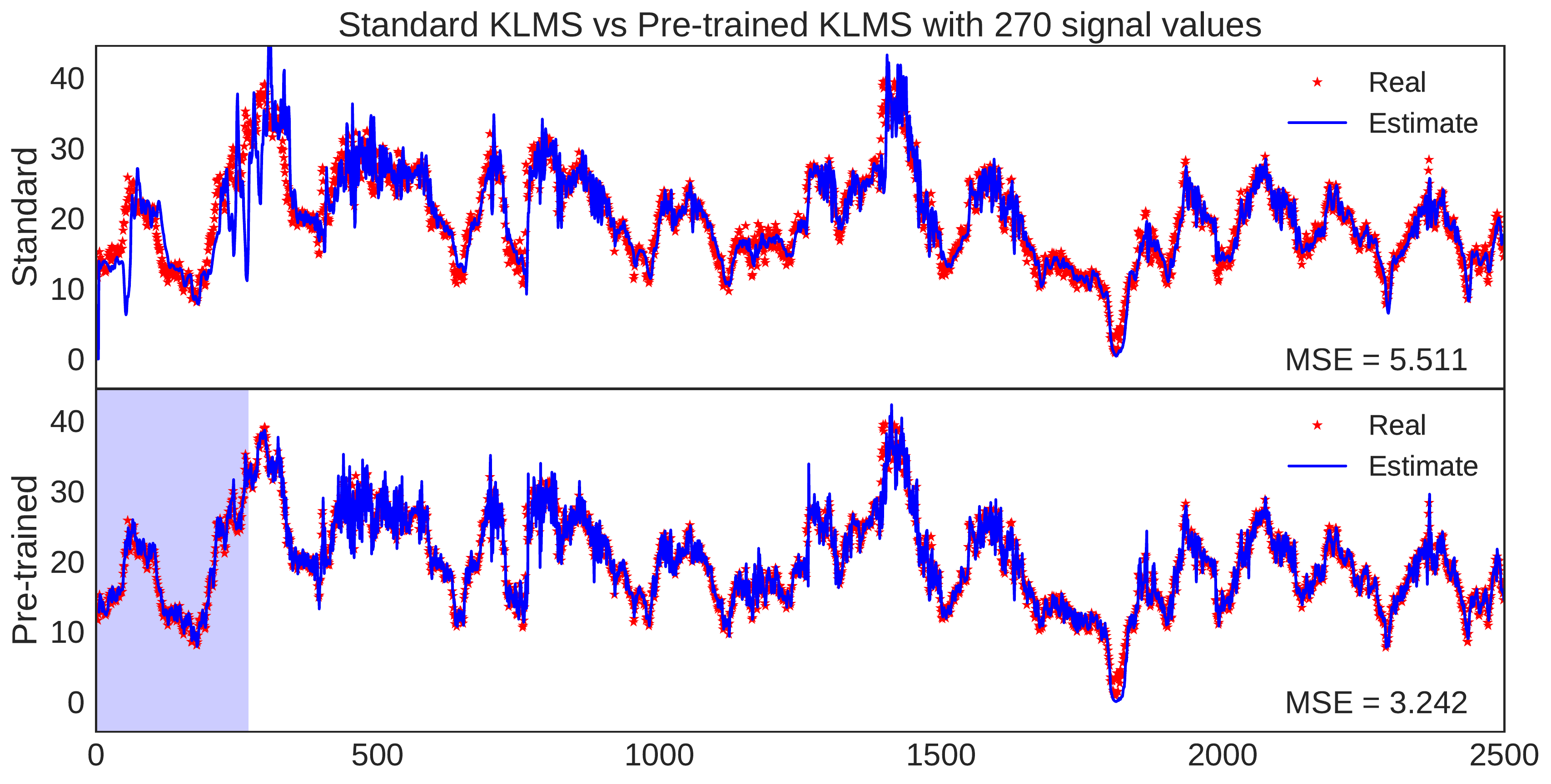}
	\caption{Wind series estimation: Standard KLMS (top) versus the proposed pre-trained KLMS with 270 training samples (bottom). Shaded area indicates training period and MSE was computed after the 270 time index for a fair comparison.}
	\label{wind-series}
\end{figure}

\begin{figure}[t!]
	\centering
	\subfloat[]{\includegraphics[width=0.17\textwidth]{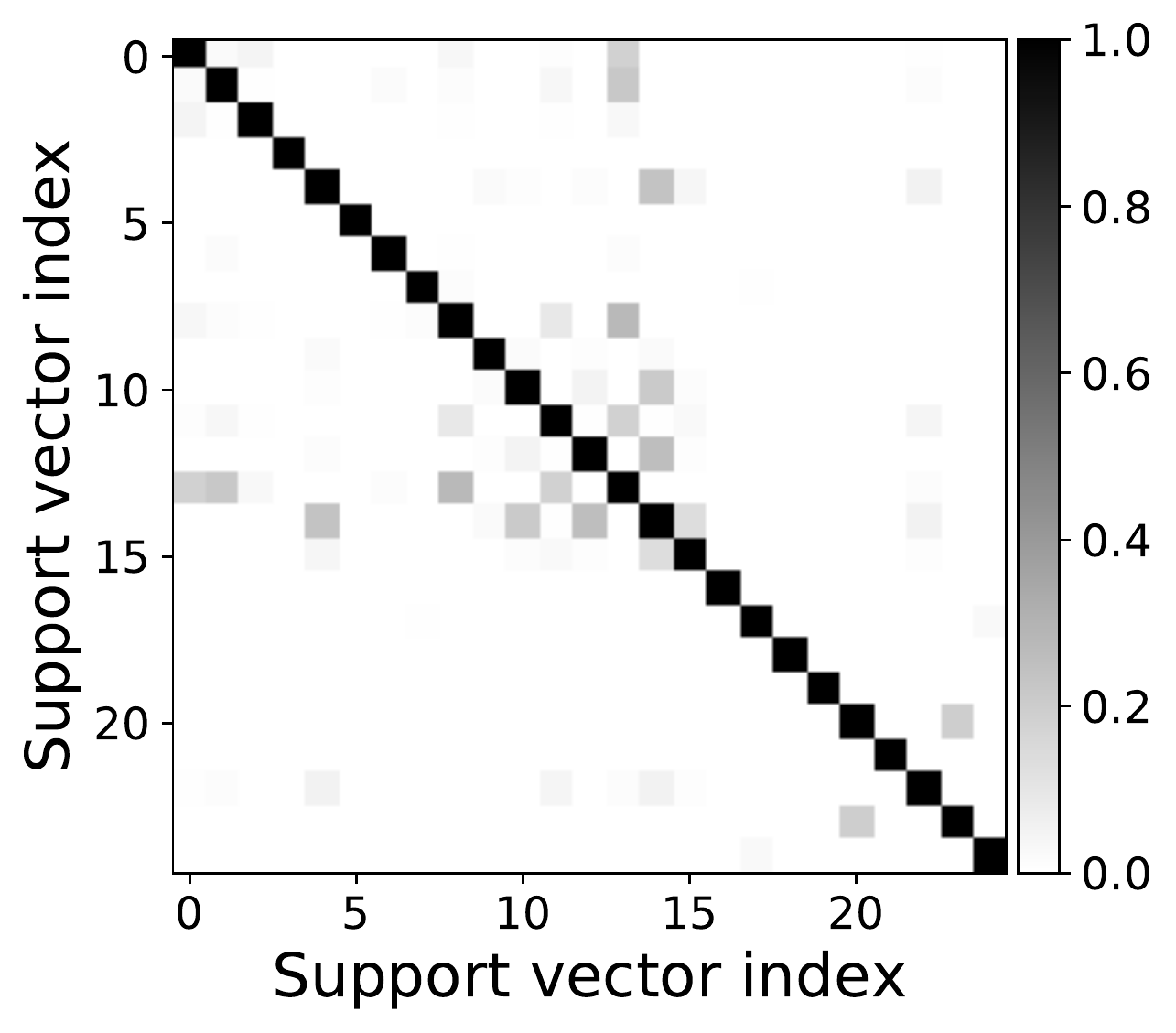}}
	\subfloat[]{\includegraphics[width=0.17\textwidth]{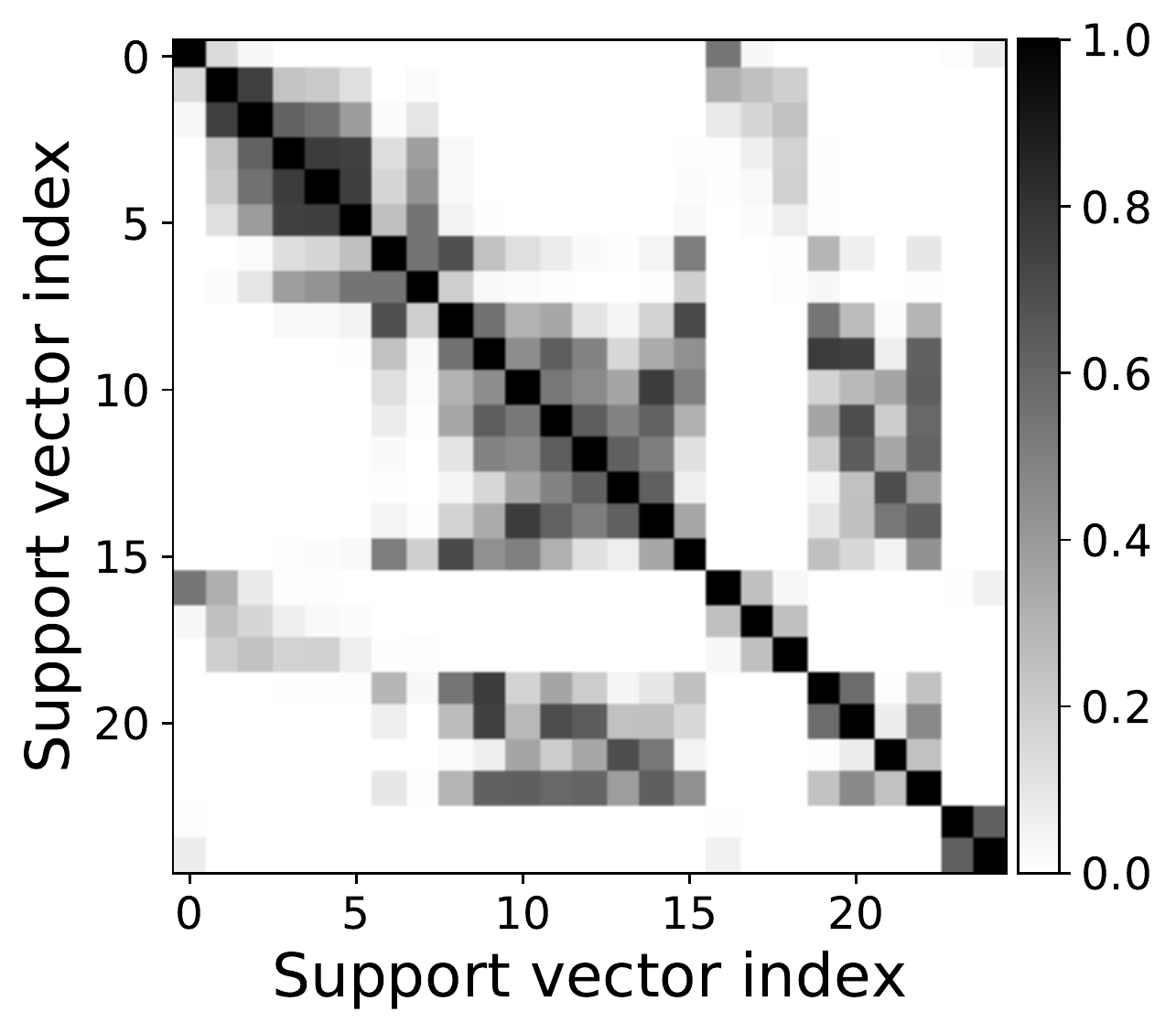}}
	\caption{Gram matrices of the pre-trained (left) and standard (right) KLMS dictionaries for wind signals. The proposed pre-trained KLMS yields a much more sparse dictionary than standard KLMS evidenced by a close-to-diagonal Gram matrix. }
	\label{wind-confmat}
\end{figure}

Fig. \ref{wind-series} shows both pre-trained KLMS and standard KLMS side by side, where the shaded area indicates the data used for pre-training. We can see that even though standard KLMS can achieve adequate performance over time, it was outperformed by the proposed pre-trained KLMS in MSE terms---{MSE was calculated after sample 270}. Furthermore, from Fig. \ref{wind-confmat} notice that the final dictionary obtained by pre-trained KLMS is much more sparse than that of standard KLMS due to the proposed sparsity-inducing prior in eq. \eqref{eq:dict_prior}, where taking the centres directly from the more than $2500$ observations resulted in centres that are too close to one another in standard KLMS.

An important finding is that, 
even though KAF methods are typically sensible to the learning rate, we found that our method is much more flexible, since desirable weights were already learned, and as such, the KLMS phase only needs a small learning rate to correct small parameter discrepancies.

\subsection{Fully-adaptive kernel adaptive filtering}

The second experiment is an online implementation of the proposed model using a sliding window. Specifically, we trained the model sequentially by (i) finding the maximum-a-posteriori (MAP) parameters  in each window, and then (ii) averaging them in time. Denoting $\theta_n^{\text{window}}$ the MAP parameters using data from the $n^\text{th}$ window, the online parameter estimate can be computed by $\theta_{n}^{\text{online}}$ according to
\begin{align}
\theta^{\text{online}}_{1} &= \theta^{\text{window}}_{1}\\ 
\theta_{n}^{\text{online}} &= \rho\theta_n^{\text{window}} + (1-\rho)\theta_{n-1}^{\text{online}}
\end{align}
where $\rho\in(0,1)$ is a forgetting factor balancing confidence between past parameter values and the new estimate. This resembles the kernel recursive least squares (KRLS) \cite{engel04,van2012kernel} rationale but rather than moving the parameters against the gradient as in KRLS, we are moving towards the MAP parameters of the current window. 

\begin{figure}[t!]
	\centering
	\includegraphics[width=0.5\textwidth]{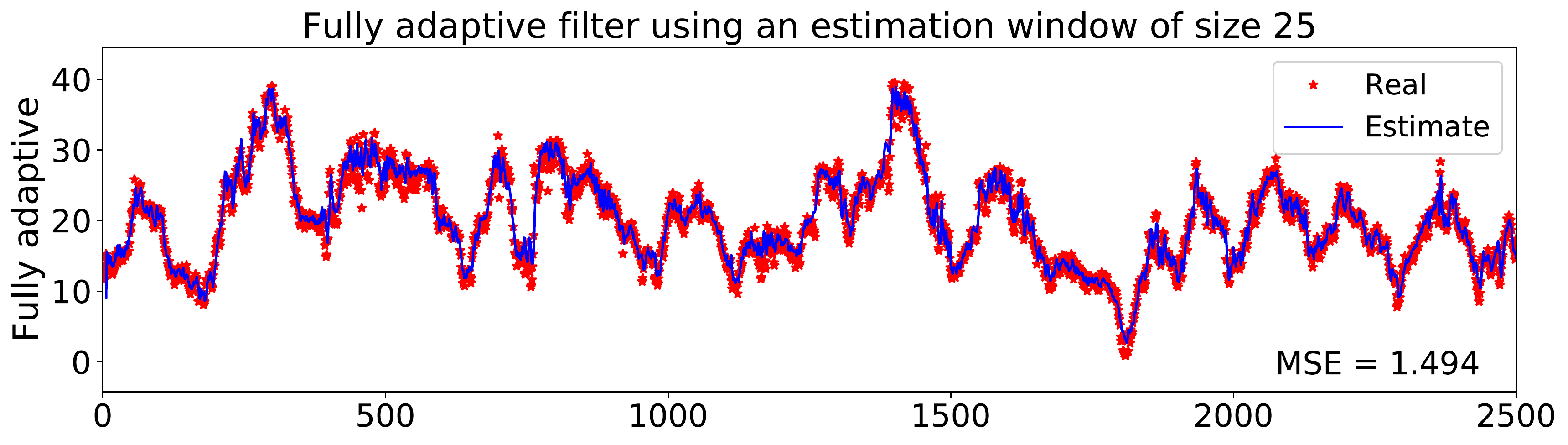}
	\caption{Fully-adaptive KAF applied to the wind series using a sliding window.}
	\label{fig:adaptive-wind}
\end{figure}

For the same wind data, we used a filter order $d=5$, a dictionary size of $10$, a forgetting factor $\rho=0.9$, a window length of $25$ samples, and $100$ MCMC samples per iteration to compute the MAP parameters. Fig. \ref{fig:adaptive-wind} shows the prediction of the proposed method trained online, where, in terms of MSE, the proposed fully-adaptive KAF implementation outperformed both the standard KLMS and the pre-trained KLMS (see the MSEs reported in Figs. \ref{wind-series} and \ref{fig:adaptive-wind}). This can be explained by the re-computation of the optimal parameters in each iteration, given a window of observations. The downside of this method is that computation time is considerable compared to standard KAF, mainly because the sampling stage of the algorithm requires compiling the log-posterior for every window before sampling even starts. We are currently improving this implementation for true online operation.

%% file: conclusions.tex
\section{Discussion and conclusions}
We have presented a probabilistic framework for pre-training kernel adaptive filters and performing fully-adaptive estimation of time series, this has been achieved by enforcing desired properties of such models via the design of meaningful prior distributions. Our pre-training approach for KAFs improves current methods both in terms of MSE and sparsity of the dictionary, thus proving that the combination of (i) the probabilistic formulation, (ii) the design of a sparsity-inducing prior, and (iii) the sample approximation of the log-posterior (MCMC) results in better prediction on both early implementation of the algorithm and future predictions. This is even clearer for patterns that were seen by the model during pre-training.

We have also showed that the proposed method can be used for fully-adaptive estimation reaching a superior
performance when compared against standard and pre-trained KAFs, and, most importantly, without hand-picking any hyperparameter, as the algorithm can effectively sample the closest combination of optimal parameters using MCMC methods. However, the complexity of the model is still an issue for online operation and alternative approaches are being developed to consolidate these promising results.